\documentclass[runningheads]{llncs}
\usepackage{graphicx}
\usepackage{comment}
\usepackage{amsmath,amssymb} % define this before the line numbering.
\usepackage{color}
\usepackage{parskip}
\usepackage[width=122mm,left=12mm,paperwidth=146mm,height=193mm,top=12mm,paperheight=217mm]{geometry}

\begin{document}
% \linenumbers
\pagestyle{headings}
\mainmatter
\def\CVPPPSubNumber{23}  % Insert your submission number here
\def\ECCVSubNumber{\CVPPPSubNumber} % needed for compatibility
\title{Usefulness of interpretability methods to explain deep learning based plant stress phenotyping} % Replace with your title

\begin{comment}
\titlerunning{CVPPP-20 submission ID \CVPPPSubNumber} 
\authorrunning{CVPPP-20 submission ID \CVPPPSubNumber} 
\author{Anonymous CVPPP submission}
\institute{Paper ID \CVPPPSubNumber}
\end{comment}
%******************

%\begin{comment}
\titlerunning{Abbreviated paper title}
\author{Koushik Nagasubramanian\ \and
Asheesh K. Singh \ \and
Arti Singh \ \and
Soumik Sarkar \ \and Baskar Ganapathysubramanian }
\authorrunning{K. Nagasubramanian et al.}
\institute{Iowa State University, Ames IA 50011, USA 
%\and
\email{\{koushikn,singhak,arti,soumiks,baskarg\}@iastate.edu}\\
}
%\end{comment}
%******************
\maketitle

\begin{abstract}
Deep learning techniques have been successfully deployed for automating plant stress identification and quantification. In recent years, there is a growing push towards training models that are interpretable - i.e. that justify their classification decisions by visually highlighting image features that were crucial for classification decisions. The expectation is that trained network models utilize image features that mimic visual cues used by plant pathologists. In this work, we compare some of the most popular interpretability methods: Saliency Maps, SmoothGrad, Guided Backpropogation, Deep Taylor Decomposition, Integrated Gradients, Layer-wise Relevance Propagation and Gradient times Input, for interpreting the deep learning model. We train a DenseNet-121 network for the classification of eight different soybean stresses (biotic and abiotic). Using a dataset consisting of 16,573 RGB images of healthy and stressed soybean leaflets captured under controlled conditions, we obtained an overall classification accuracy of 95.05 \%. For a diverse subset of the test data, we compared the important features with those identified by a human expert. We observed that most interpretability methods identify the infected regions of the leaf as important features for some -- but not all -- of the correctly classified images. For some images, the output of the interpretability methods indicated that spurious feature correlations may have been used to correctly classify them. Although the output explanation maps of these interpretability methods may be different from each other for a given image, we advocate the use of these interpretability methods as  `hypothesis generation' mechanisms that can drive scientific insight.
\end{abstract}

\section{Introduction}
Convolutional Neural Networks (CNN) based Deep Learning (DL) architectures have led to rapid developments in diverse applications such as object recognition, document reading and sentiment analysis within the industry and social media \cite{krizhevsky2012imagenet}. There is now an increasing push to leverage CNN’s for science applications with several exciting advances in the context of plant phenotyping \cite{pound2017deep}. A particularly promising application area within plant phenotyping is the utilization of CNNs for large scale plant stress phenotyping, which has the potential to transform disease scouting, crop management and breeding for climate change  \cite{singh2018deep}. From the scientific standpoint, it is increasingly becoming clear that having a predictive model alone is not enough, and there is an interest in qualitative (if not quantitative) measures that could justify trusting the decision making of the model, as well as the need to extract scientific insight from the predictive model. One recent thrust has been to design so-called interpretability methods \cite{xie2020explainable} that seek to visualize the key latent features used by the model for making predictions. Essentially, if these key visualized features make physiological sense, then the premise is that model is making the right decision for the correct reasons. This is akin to a plant pathologist explaining her diagnosis of a plant stress by pointing out key visual cues used to arrive at their decision. We explore the utility of various interpretability methods for an important science application in plant pathology. 

\noindent We specifically focus on interpretability of CNNs that have been successfully deployed for plant disease detection. This is important because plant diseases negatively impact yield potential of crops worldwide, including soybean [Glycine max (L.) Merr.], reducing the average annual soybean yield by an estimated 11\% in the United States \cite{hartman2015compendium}. However, currently disease scouting and phenotyping techniques predominantly rely on human scouts and visual ratings. Human visual ratings are dependent on rater ability, rater reliability, and can be prone to human error, subjectivity, and inter/intra-rater variation \cite{Bock2010}. This motivates the need for improved technologies for disease detection and identification beyond visual ratings and in an earlier timescale to improve yield protection, disease mitigation and reduce significant waste of resources.  We trained a CNN model to effectively classify a diverse set of foliar stresses in soybean from red, green, blue (RGB) images of soybean leaflets \cite{ghosal2018explainable}. We reliably classified several biotic (bacterial and fungal diseases) and abiotic (chemical injury and nutrient deficiency) stresses by learning from over 16,500 images. However, limited information is available that compares currently available interpretability models in the application to plant disease phenotyping problem. The interpretability focus for DL based plant phenotyping is due to practitioner’s requirement to move away from a black-box predictor and remove the hesitancy of representative end-user to believe DL predictions. This links or bridges the gap between visual symptom-based manual identification (by trained plant pathologists) that include an explanation mechanism, for example, visible chlorosis and necrosis are symptomatic of Iron Deficiency Chlorosis (IDC), and DL method outputs. The trained model can be deployed as a mobile application for real-time detection of soybean stresses by farmers and researchers.\par
\noindent Previous works on interpretability mechanisms applied to plant science problems have predominately focused on the advantages of such methods \cite{sladojevic2016deep,ballester2017assessing,brahimi2017deep,brahimi2018deep,brahimi2019deep,nagasubramanian2019plant}. More recently, Toda et al. \cite{toda2019convolutional} compared different interpretability methods for identifying the important features used in plant disease recognition and discussed some of its advantages and limitations. They suggested that interpretability methods could potentially be used to identify the reasons for misclassification and dataset bias. They also used these visualization methods for pruning a deep learning model.\par

\noindent Here, in addition to critically evaluating some of the commonly used interpretability methods for identifying feature importance, we make some practical observations on the utility of interpretability methods to plant stress phenotyping. This includes a) interpretability methods can enable identification of spurious correlations due to dataset bias even for correctly classified images and b) even misclassified images can have visually good interpretability outputs indicating the possibility of confounding symptoms. These observations indicate that even images with good interpretability visualizations can be misclassified and some images with visually poor interpretability outputs could be correctly classified. While generally useful, outputs of these explanation methods can visually look different from each other for a given image. While this does not preclude using interpretability methods for plant stress phenotyping, we advocate the use of these methods as ‘hypothesis generation’ mechanisms to identify novel visual cues as potential disease signatures. Therefore, interpretability methods can be used to test model trustworthiness, and produce potential hypotheses that can then be rigorously scientifically evaluated.

% \noindent Recent interpretability works on plant science problems have mainly focused on the advantages of those methods . In this work, we evaluate the usefulness and limitations of some of the most popular local explanation methods for the identification of important input features used for a particular classification decision. While generally useful, our results identified that outputs of these explanation methods could visually look different from each other for a given image. While this does not preclude using interpretability methods for plant stress phenotyping, we advocate the use of these methods as ‘hypothesis generation’ mechanisms to identify novel visual cues as potential disease signatures. Therefore, interpretability methods can be used to test model trustworthiness, and produce potential hypotheses that can then be rigorously scientifically evaluated.
\section{Materials and Methods}
\subsection{Dataset}
\begin{figure}
\centering
\includegraphics[height=6.5cm,width=12cm]{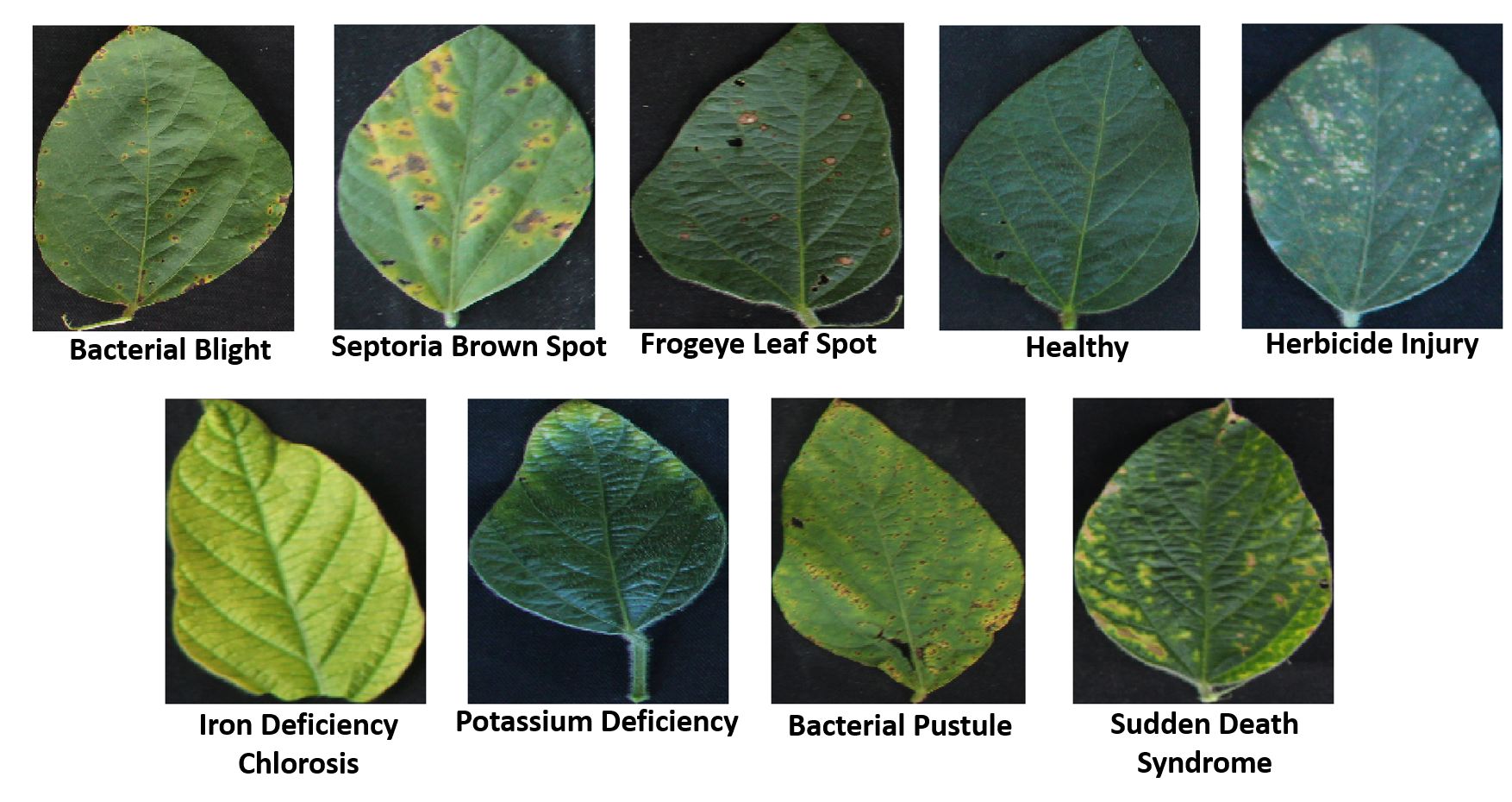}
\caption{Image examples of healthy leaflet and 8 different soybean stresses}
\label{fig:soy_c}
\end{figure}

The dataset consists of 16573 RGB images of soybean leaflets across 9 different classes (8 different soybean stresses and healthy soybean leaflets). These classes cover a diverse spectrum of biotic and abiotic foliar stresses. Fig.~\ref{fig:soy_c} illustrates the imaging setup and the 9 different soybean stress classes used in our study. The training and test data consisted of 14915 (90 \%) and 1658 (10 \%) images, respectively. The data split across the 9 classes is shown in Table~\ref{table:classes}. Please refer to \cite{ghosal2018explainable} for more details on the dataset collection.
\setlength{\tabcolsep}{4pt}
\begin{table}
\begin{center}
\caption{Composition of training and test data split among the 9 different classes for soybean healthy and stresses (biotic and abiotic).}
\label{table:classes}
\begin{tabular}{lll}
\hline\noalign{\smallskip}
Class & Training & Testing\\
\noalign{\smallskip}
\hline
\noalign{\smallskip}
Healthy  & 3800 & 423\\
Bacterial Blight & 1372 & 152\\
Brown Spot & 1222 & 136\\
Frog-eye leaf spot & 1010 & 112\\
Herbicide Injury & 1255 & 140\\
IDC & 1660 & 184 \\
Potassium Deficiency & 1967 & 219\\
Bacterial Pustule & 1507 & 167\\
SDS & 1122 & 125\\
\hline
\end{tabular}
\end{center}
\end{table}
\subsection{Model architecture and training}
\begin{figure}
\centering
\includegraphics[height=3.5cm,width=12cm]{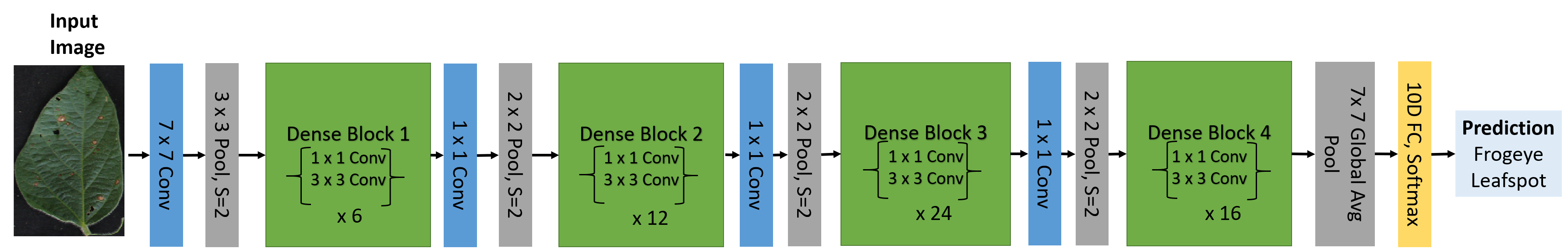}
\caption{Illustration of the Densenet-121 network architecture. Conv represents Batch Normalization-Convolution-ReLU layers, Pool represents max-pooling layer and S represents the stride value. For representation, Frogeye leaf spot example is included in the input. }
\label{fig:dense}
\end{figure}

We used DenseNet-121 architecture for the classification of soybean stress diseases. DenseNet iteratively concatenates the feature maps from one layer to another layer along the \cite{Huang2016}, which has been shown to be useful for classification tasks. Fig.~\ref{fig:dense} shows the illustration of DenseNet-121 architecture. The DenseNet-121 model architecture was chosen because of its better performance over MobileNetV2 \cite{sandler2018mobilenetv2}, ResNet50 \cite{he2016deep} and NASNet \cite{zoph2018learning} architectures. We randomly chose 10\% of the training data for the validation set in each epoch. The CNN model was trained for 200 epochs and early stopping method was used to find the best model on the validation dataset . We trained using Keras \cite{chollet2015keras} with Tensorflow \cite{abadi2016tensorflow}. Adam optimizer was used to train our model on mini-batches of size 32 \cite{kingma2014adam} with a learning rate value of $1 \times 10^{-3}$. Training was performed on a NVIDIA Tesla P40 GPU. 
\subsection{Model Interpretation}
Given an input image and the class predicted by the trained CNN model for this input image, interpretability methods compute an importance score for each pixel in the input image. This importance image is normalized and visualized to see if the most important cluster of pixels in fact have any physiological meaning. Formally, we have a function $\boldsymbol{F: \mathbb{R}^d \rightarrow \mathbb{R}^c}$  that represents a neural network where $\boldsymbol{c}$ is the number of output classes. In our setup, $\boldsymbol{x \in \mathbb{R}^d}$ is an input vector to the neural network model and $\boldsymbol{A_n^l}$ is the output activation of neuron $\boldsymbol{n}$ in layer $\boldsymbol{l}$. The interpretability method provides an output explanation map $\boldsymbol{E}$, which has the importance score for each pixel in the input image. In this work, we choose the layer $\boldsymbol{l}$ to be the fully connected layer before the output softmax layer and the neuron $\boldsymbol{n}$ is the one with maximum output activation in layer $\boldsymbol{l}$. We provide a brief overview of each method, with links to references for the interested reader. These interpretability methods were implemented using the Keras based iNNvestigate library \cite{alber2019innvestigate}.
\subsubsection{Saliency map.}
It is gradient of the unnormalized output class activation with respect to the input \cite{simonyan2013deep}.
\begin{align}
    \boldsymbol{E = \frac{\partial A_n^l}{\partial x}}
\end{align}
\subsubsection{Gradient times Input (GI).}
It uses element-wise product of the input and the gradient to overcome gradient saturation.
\begin{align}
    \boldsymbol{E = x \odot \frac{\partial A_n^l}{\partial x}}
\end{align}
\subsubsection{Guided Backpropagation (GBP)}
It uses a modified backpropagation for explanation by setting the negative gradient values to zero while backpropagating through a ReLU unit \cite{Springenberg2014}
\subsubsection{SmoothGRad (SG).}
It reduces the noisy output of the saliency map $\boldsymbol{E}$ by averaging over gradient explanations of noisy variations of an input \cite{smilkov2017smoothgrad}.
\begin{align}
    \boldsymbol{Esg = \frac{1}{N}\sum_{i=1}^{N}E(x + n_i)}
\end{align}
where $n_i$ represents the gaussian noise added to the input image.
\subsubsection{Integrated Gradients (IG).}
It overcomes gradient saturation by summing over explanations over scaled versions of the input \cite{Sundararajan2017}.
\begin{align}
    \boldsymbol{E = (x - \bar{x}) \times \int_{0}^{1}\frac{\partial F(\bar{x} + \alpha (x - \bar{x}}{\partial x}\partial \alpha}
\end{align}
\subsubsection{Deep Taylor Decomposition (DTD).}
It decomposes the prediction of a deep neural network down to relevance scores of the input features using Taylor expansion of a function at some chosen point \cite{montavon2017explaining}.
\subsubsection{Layerwise Relevance Propagation (LRP).}
These LRP-z rule decompose the prediction of a deep neural network down to relevance scores of the input features using local redistribution rules based on weighted neuron activation (z) \cite{Bach2015}. Recently, \cite{kindermans2019reliability} have shown that Gradient times input method is equivalent to LRP-z for networks with ReLU activation.
\section{Results and Discussion}
\begin{figure}[h]
\centering
\includegraphics[height=10cm,width=10cm]{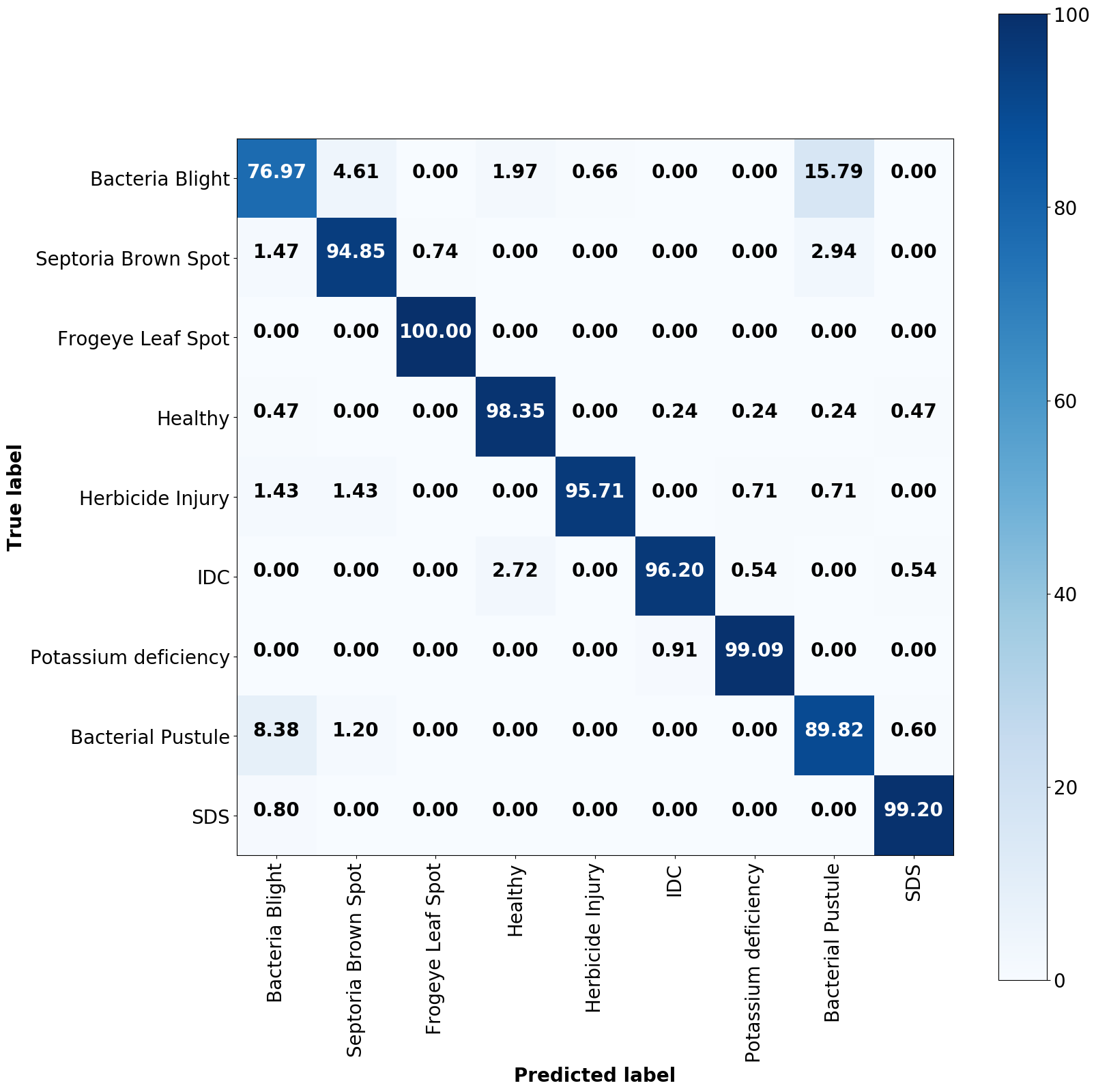}
\caption{Classification accuracy confusion matrix from eight soybean stresses and healthy classes. }
\label{fig:dense_acc}
\end{figure}
\subsection{Classification}
The overall 9 class classification accuracy of the DenseNet-121 model was 95.05\%. Fig.~\ref{fig:dense_acc} illustrates the normalized confusion matrix of the classification. The confusion matrix revealed that erroneous predictions were predominantly due to confounding stress symptoms. Specifically, the highest classification error occurred between bacterial blight and bacterial pustule. Erroneously, 15.79\% of bacterial pustule test images were predicted as bacterial blight, and 8.38\% of bacterial blight test images were predicted as bacterial pustule. Discriminating between these two diseases is challenging even for expert plant pathologists due to their confounding stress symptoms.

\subsection{Interpretability evaluation}
A trained plant pathologist evaluated the interpretability outputs. A local explanation output is considered good if the pixels that were identified as important by the interpretability method was also identified as important visual cues by the plant pathologist.We make several observations from this exercise:\par
\begin{figure}
\centering
\includegraphics[height=12cm,width=12cm]{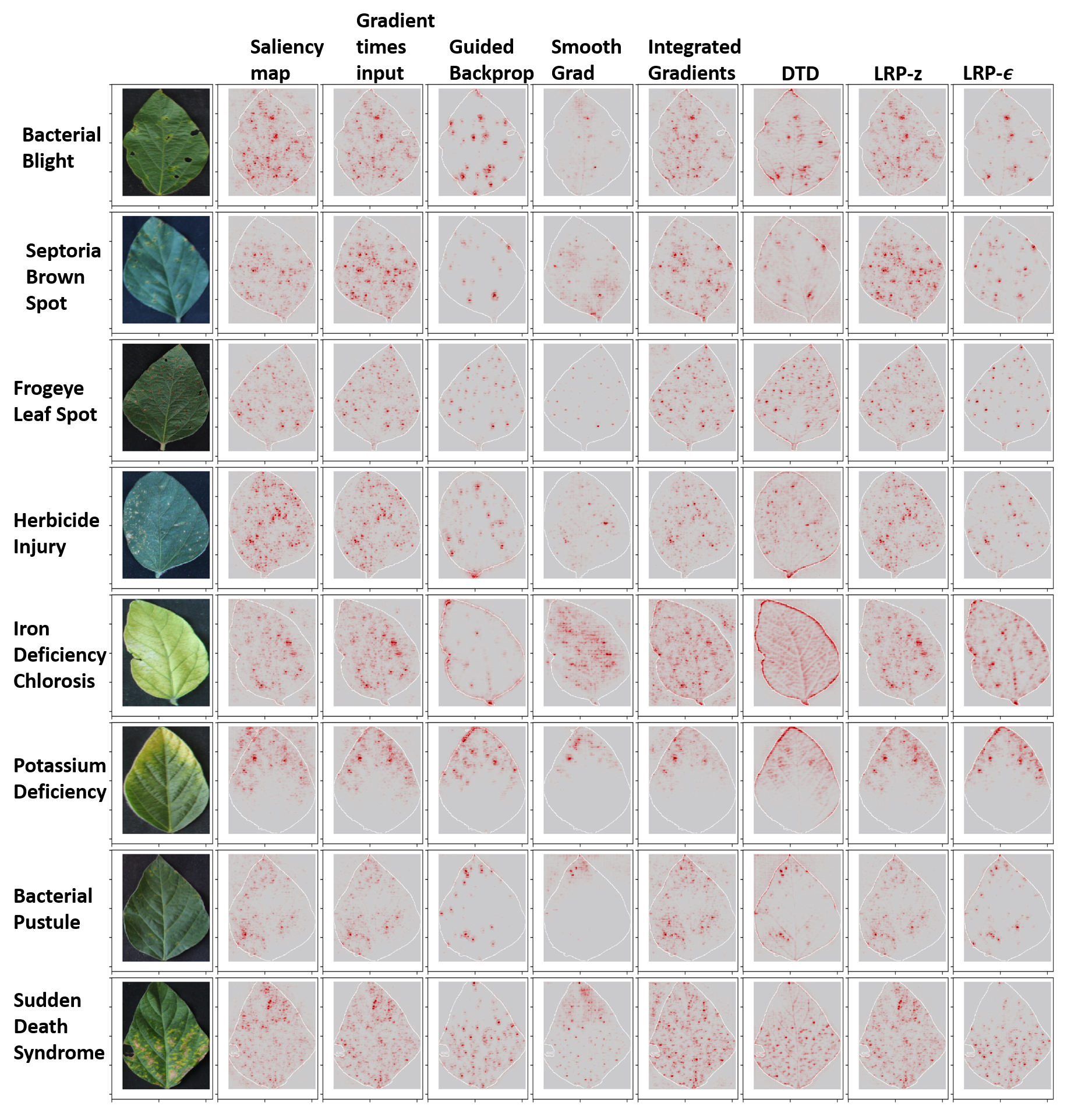}
\caption{Evaluation of local explanation methods on DenseNet-121 model for 8 different stress classes of input images which were classified correctly. The explanation outputs were normalized in the range of [0,1]. The outer edge of the leaf is shown in white color. On visual inspection, we observed that at least some part of the infected region in each image were identified as important features (dark red color) by the explanation methods.}
\label{fig:vis_g}
\end{figure}
\noindent \textbf{Observation 1:} A large fraction of images that were correctly classified by the CNN model produced local explanation outputs that matched the visual cues identified by the plant pathologist. Fig.~\ref{fig:vis_g} illustrates some of the good local explanation outputs for the 9 different classes of images which were classified correctly. We observed this to be particularly the case when localized symptoms determined the disease stress. That is, diseases like Septoria Brown Spot, Frogeye Leaf Spot and Bacterial Pustule, produce sparse and localized symptoms that are correctly identified by the interpretability methods. For these sparse symptoms of the disease on leaflet, the explanations from all the interpretability methods assigned high importance to infected pixel locations containing visual symptoms and low attributions to healthy pixel locations of the leaf as shown in Fig.~\ref{fig:vis_g}. For abiotic stress such as Iron Deficiency Chlorosis where symptoms are spread throughout the leaf area, the explanations from all interpretability methods indicate that the models use pixel locations from only some part of the infected leaf area for the classification decision instead of the complete infected leaf area. These sub-features could be visual symptoms worthy of additional investigation by plant pathologists. On visual inspection, the saliency map method appears noisy and scattered compared to other methods. Across all methods, some of the infected pixels are consistently indicated as important features across different interpretability methods.\par 

\begin{figure}
\centering
\includegraphics[height=12cm,width=12cm]{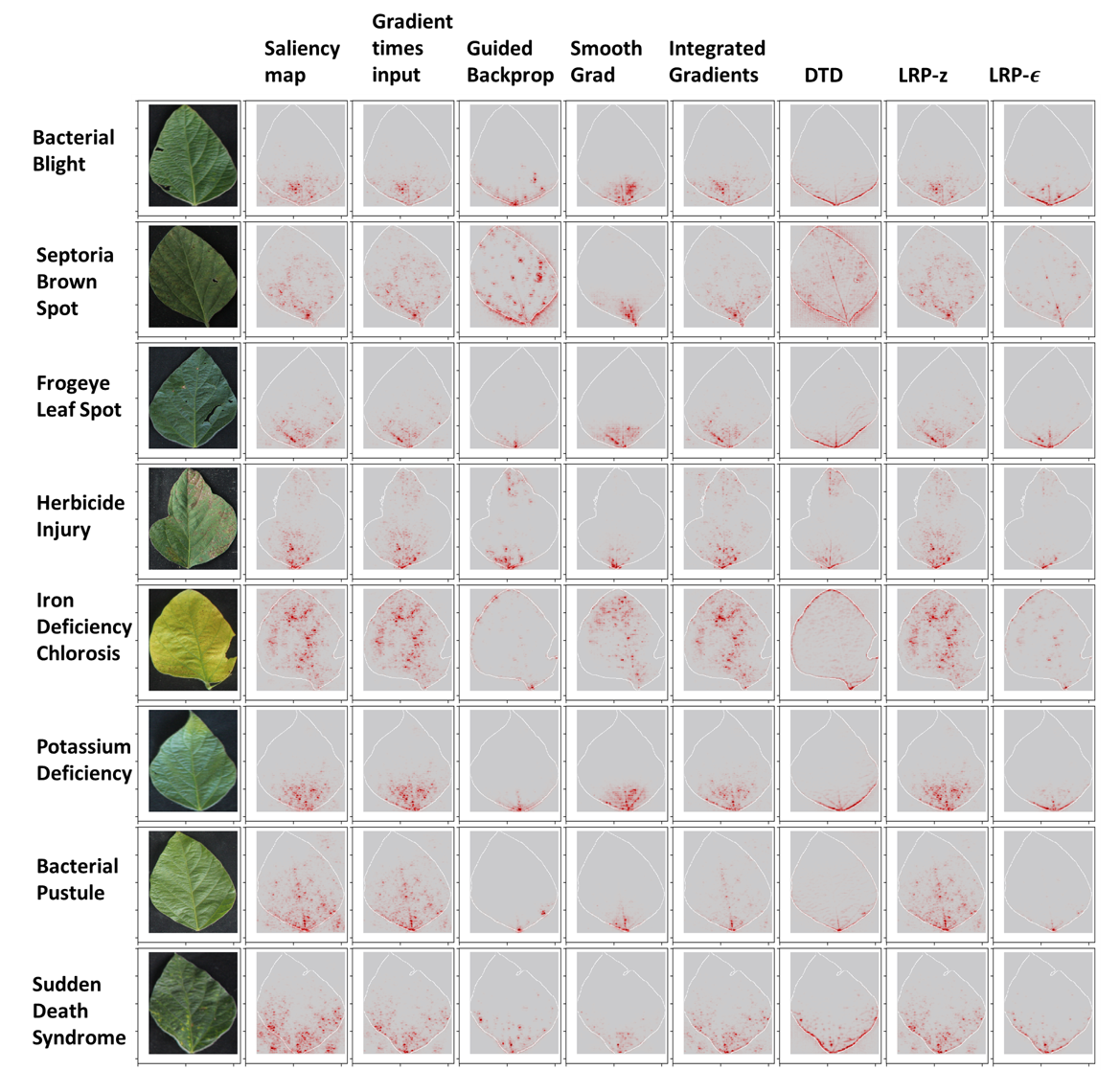}
\caption{Evaluation of explanation methods on DenseNet-121 model for 8 different classes of input images which were classified correctly. The interpretability outputs were normalized in the range of [0,1]. The outer edge of the leaf is shown in white color. On visual inspection, we observed that the explanation outputs were not localized on the highly infected regions even though the images were classified correctly.}
\label{fig:vis_b_cc}
\end{figure}

\noindent \textbf{Observation 2:} While majority of the explanations were accurate, some images that were correctly classified by the CNN model produced poor explanation outputs. Fig.~\ref{fig:vis_b_cc} illustrates some of the poor local explanation outputs for the 9 different classes of images which were correctly classified. We observed that local explanation outputs did not match the pathologist visual cues for stress like Iron Deficiency Chlorosis, where symptoms are spread throughout the leaf area. Either the explanations outputs are limited to only some part of the infected leaf area or were focused on the leaf edge where this is a color gradient. Importantly, this indicates that a poor local explanation output does not necessarily indicate misclassification. Moreover, this could also indicate that the CNN is using some spurious feature correlations with individual diseases for making the classification decisions due to dataset bias \cite{geirhos2020shortcut} On visual inspection of the output of interpretability methods for randomly chosen 90 images (10 images per class), we observe that 6 (5 images from Frogeye Leaf Spot and 1 image from potassium deficiency) used spurious correlations for the classification decision similar to examples shown in Fig.~\ref{fig:vis_b_cc}.\par
\begin{figure}
\centering
\includegraphics[height=12cm,width=12cm]{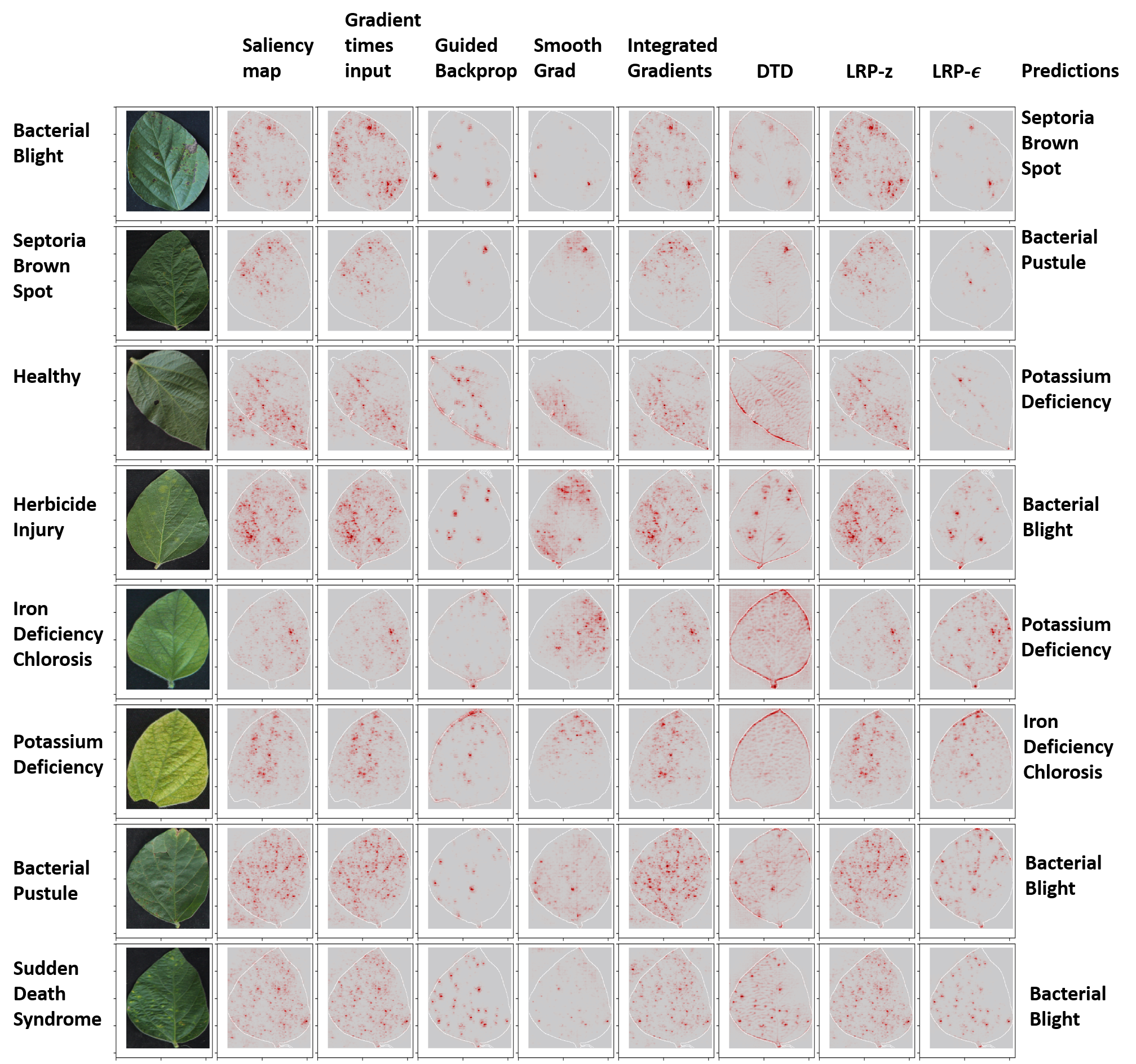}
\caption{Evaluation of explanation methods on DenseNet-121 model for 8 different classes of input images which were mis-classified. The interpretability outputs were normalized in the range of [0,1]. The outer edge of the leaf is shown in white color. On visual inspection, we observed that the explanation outputs were localized on some part of the infected regions even though the images were misclassified.}
\label{fig:vis_mis}
\end{figure}

\noindent \textbf{Observation 3:} Even images that were misclassified by the CNN model produced local explanation outputs that matched the visual cues identified by the plant pathologist. Fig.~\ref{fig:vis_mis}. illustrates that the explanation outputs could be good even for some of the misclassified images. This was particularly apparent for disease symptoms of one class that can be confounded with another class. For example, the disease symptoms of Bacterial Blight could have been confounded with Bacterial Pustule as shown in Fig.~\ref{fig:vis_mis}. In fact, these confounding visual symptoms make correct classification a challenge even for the human expert.\par
\noindent Overall, in spite of some visual differences between the  outputs of different interpretability methods for a given image, we emphasize  that these interpretability methods can be very useful to (a) identify and extend the suite of current visual cues used by pathologists for disease classification (given Observation 1), (b) identify if spurious features were used for classification decision (given Observation 2) and (c) evaluate and quantify impact of confounding symptoms (given Observation 3). We call such a deployment of interpretability methods as a ‘hypothesis generation’ tool, and the success in local explanation outputs from interpretability models encourages utilization of explainable DL to become more mainstream in plant phenotyping applications.
\section{Conclusions}
We have demonstrated that a DenseNet-121 model can effectively be used to classify 8 different stresses in soybeans. We used some of the commonly used interpretability methods to identify the pixels that were important for the classification. The saliency map, GI, GBP, SG, IG, DTD, LRP-z and LRP-$\epsilon$ identified the pixels that influence the classification score.\par

\noindent We conclude with several suggestions: (a) Interpretability methods are useful but have to be carefully and appropriately vetted and utilized, (b) All these local explanation methods are based on first order approximation of the neural network and potentially suffer from the significant nonlinearity of the neural network models \cite{samek2019explainable}. We anticipate rapid progress in the development of more sophisticated interpretability methods that could potentially tease out causal relationships between symptoms and classification decisions for highly nonlinear neural network models \cite{du2019techniques}, (c) until such tools are available, we encourage the plant science community to utilize interpretability methods as ‘hypothesis generation’ tools to provide feedback to the domain scientists. 
\section{Acknowledgments}
We thank D. Blystone, J. Brungardt, B. Scott, and undergraduate students (A.K.S. laboratory) for their help in imaging and data collection.

\section{Author contributions} KN, AS, AKS, SS, BG formulated the problem and designed the study; KN ran numerical experiments; KN analyzed data with help from all authors; All authors contributed to writing the paper. 

\section{Funding}
This work was funded by the Iowa Soybean Association (A.S.); an Iowa State University (ISU) internal grant (to all authors); a NSF/USDA National Institute of Food and Agriculture grant (to all authors); a Monsanto Chair in Soybean Breeding at Iowa State University (A.K.S.); Raymond F. Baker Center for Plant Breeding at Iowa State University (A.K.S.); an ISU Plant Sciences Institute fellowship (to B.G., A.K.S., and S.S.), and USDA IOW04314 (to A.S. and A.K.S.).

\section{Competing interests} 
The author(s) declare(s) that there is no conflict of interest regarding the publication of this article.

\section{Data and Code Availability}
The data and code will be made publicly available soon.

\clearpage
\bibliographystyle{splncs04}
\bibliography{inter_r}

\begin{thebibliography}{10}
\providecommand{\url}[1]{\texttt{#1}}
\providecommand{\urlprefix}{URL }
\providecommand{\doi}[1]{https://doi.org/#1}

\bibitem{abadi2016tensorflow}
Abadi, M., Barham, P., Chen, J., Chen, Z., Davis, A., Dean, J., Devin, M.,
  Ghemawat, S., Irving, G., Isard, M., Others: {TensorFlow: A System for
  Large-Scale Machine Learning.} In: Osdi. vol.~16, pp. 265--283 (2016).
  \doi{10.1038/nn.3331}

\bibitem{alber2019innvestigate}
Alber, M., Lapuschkin, S., Seegerer, P., H{\"a}gele, M., Sch{\"u}tt, K.T.,
  Montavon, G., Samek, W., M{\"u}ller, K.R., D{\"a}hne, S., Kindermans, P.J.:
  innvestigate neural networks. Journal of Machine Learning Research
  \textbf{20}(93), ~1--8 (2019)

\bibitem{Bach2015}
Bach, S., Binder, A., Montavon, G., Klauschen, F., M{\"{u}}ller, K.R., Samek,
  W.: {On pixel-wise explanations for non-linear classifier decisions by
  layer-wise relevance propagation}. PLoS ONE  \textbf{10}(7),  e0130140 (jul
  2015). \doi{10.1371/journal.pone.0130140}

\bibitem{ballester2017assessing}
Ballester, P., Correa, U.B., Birck, M., Araujo, R.: {Assessing the performance
  of convolutional neural networks on classifying disorders in apple tree
  leaves}. In: Latin American Workshop on Computational Neuroscience. pp.
  31--38. Springer (2017)

\bibitem{Bock2010}
Bock, C.H., Poole, G.H., Parker, P.E., Gottwald, T.R.: {Plant disease severity
  estimated visually, by digital photography and image analysis, and by
  hyperspectral imaging}. Critical Reviews in Plant Sciences  \textbf{29}(2),
  59--107 (2010). \doi{10.1080/07352681003617285}

\bibitem{brahimi2018deep}
Brahimi, M., Arsenovic, M., Laraba, S., Sladojevic, S., Boukhalfa, K.,
  Moussaoui, A.: {Deep learning for plant diseases: detection and saliency map
  visualisation}. In: Human and Machine Learning, pp. 93--117. Springer (2018)

\bibitem{brahimi2017deep}
Brahimi, M., Boukhalfa, K., Moussaoui, A.: {Deep learning for tomato diseases:
  classification and symptoms visualization}. Applied Artificial Intelligence
  \textbf{31}(4),  299--315 (2017)

\bibitem{brahimi2019deep}
Brahimi, M., Mahmoudi, S., Boukhalfa, K., Moussaoui, A.: {Deep interpretable
  architecture for plant diseases classification}. In: 2019 Signal Processing:
  Algorithms, Architectures, Arrangements, and Applications (SPA). pp.
  111--116. IEEE (2019)

\bibitem{chollet2015keras}
Chollet, F., Others: {Keras}. $\backslash$url{\{}https://keras.io{\}} (2015),
  \url{https://keras.io}

\bibitem{du2019techniques}
Du, M., Liu, N., Hu, X.: Techniques for interpretable machine learning.
  Communications of the ACM  \textbf{63}(1),  68--77 (2019)

\bibitem{geirhos2020shortcut}
Geirhos, R., Jacobsen, J.H., Michaelis, C., Zemel, R., Brendel, W., Bethge, M.,
  Wichmann, F.A.: {Shortcut Learning in Deep Neural Networks}. arXiv preprint
  arXiv:2004.07780  (2020)

\bibitem{ghosal2018explainable}
Ghosal, S., Blystone, D., Singh, A.K., Ganapathysubramanian, B., Singh, A.,
  Sarkar, S.: {An explainable deep machine vision framework for plant stress
  phenotyping}. Proceedings of the National Academy of Sciences
  \textbf{115}(18),  4613--4618 (2018)

\bibitem{hartman2015compendium}
Hartman, G.L., Rupe, J.C., Sikora, E.F., Domier, L.L., Davis, J.A., Steffey,
  K.L.: {Compendium of Soybean Diseases and Pests.} Am Phytopath Society (2015)

\bibitem{he2016deep}
He, K., Zhang, X., Ren, S., Sun, J.: {Deep residual learning for image
  recognition}. In: Proceedings of the IEEE conference on computer vision and
  pattern recognition. pp. 770--778 (2016)

\bibitem{Huang2016}
Huang, G., Liu, Z., van~der Maaten, L., Weinberger, K.Q.: {Densely Connected
  Convolutional Networks}. Proceedings - 30th IEEE Conference on Computer
  Vision and Pattern Recognition, CVPR 2017  \textbf{2017-January},  2261--2269
  (aug 2016), \url{http://arxiv.org/abs/1608.06993}

\bibitem{kindermans2019reliability}
Kindermans, P.J., Hooker, S., Adebayo, J., Alber, M., Sch{\"{u}}tt, K.T.,
  D{\"{a}}hne, S., Erhan, D., Kim, B.: {The (un) reliability of saliency
  methods}. In: Explainable AI: Interpreting, Explaining and Visualizing Deep
  Learning, pp. 267--280. Springer (2019)

\bibitem{kingma2014adam}
Kingma, D., Ba, J.: {Adam: A method for stochastic optimization}. arXiv
  preprint arXiv:1412.6980  (2014)

\bibitem{krizhevsky2012imagenet}
Krizhevsky, A., Sutskever, I., Hinton, G.E.: {Imagenet classification with deep
  convolutional neural networks}. In: Advances in neural information processing
  systems. pp. 1097--1105 (2012)

\bibitem{montavon2017explaining}
Montavon, G., Lapuschkin, S., Binder, A., Samek, W., M{\"u}ller, K.R.:
  Explaining nonlinear classification decisions with deep taylor decomposition.
  Pattern Recognition  \textbf{65},  211--222 (2017)

\bibitem{nagasubramanian2019plant}
Nagasubramanian, K., Jones, S., Singh, A.K., Sarkar, S., Singh, A.,
  Ganapathysubramanian, B.: {Plant disease identification using explainable 3D
  deep learning on hyperspectral images}. Plant methods  \textbf{15}(1), ~98
  (2019)

\bibitem{pound2017deep}
Pound, M.P., Atkinson, J.A., Townsend, A.J., Wilson, M.H., Griffiths, M.,
  Jackson, A.S., Bulat, A., Tzimiropoulos, G., Wells, D.M., Murchie, E.H.,
  Others: {Deep machine learning provides state-of-the-art performance in
  image-based plant phenotyping}. Gigascience  \textbf{6}(10),  gix083 (2017)

\bibitem{samek2019explainable}
Samek, W.: {Explainable AI: interpreting, explaining and visualizing deep
  learning}, vol. 11700. Springer Nature (2019)

\bibitem{sandler2018mobilenetv2}
Sandler, M., Howard, A., Zhu, M., Zhmoginov, A., Chen, L.C.: {Mobilenetv2:
  Inverted residuals and linear bottlenecks}. In: Proceedings of the IEEE
  conference on computer vision and pattern recognition. pp. 4510--4520 (2018)

\bibitem{simonyan2013deep}
Simonyan, K., Vedaldi, A., Zisserman, A.: {Deep inside convolutional networks:
  Visualising image classification models and saliency maps}. arXiv preprint
  arXiv:1312.6034  (2013)

\bibitem{singh2018deep}
Singh, A.K., Ganapathysubramanian, B., Sarkar, S., Singh, A.: {Deep learning
  for plant stress phenotyping: trends and future perspectives}. Trends in
  plant science  \textbf{23},  883--898 (2018)

\bibitem{sladojevic2016deep}
Sladojevic, S., Arsenovic, M., Anderla, A., Culibrk, D., Stefanovic, D.: {Deep
  neural networks based recognition of plant diseases by leaf image
  classification}. Computational intelligence and neuroscience  \textbf{2016}
  (2016)

\bibitem{smilkov2017smoothgrad}
Smilkov, D., Thorat, N., Kim, B., Vi{\'e}gas, F., Wattenberg, M.: Smoothgrad:
  removing noise by adding noise. arXiv preprint arXiv:1706.03825  (2017)

\bibitem{Springenberg2014}
Springenberg, J.T., Dosovitskiy, A., Brox, T., Riedmiller, M.: {Striving for
  Simplicity: The All Convolutional Net}. 3rd International Conference on
  Learning Representations, ICLR 2015 - Workshop Track Proceedings  (dec 2014),
  \url{http://arxiv.org/abs/1412.6806}

\bibitem{Sundararajan2017}
Sundararajan, M., Taly, A., Yan, Q.: {Axiomatic Attribution for Deep Networks}.
  34th International Conference on Machine Learning, ICML 2017  \textbf{7},
  5109--5118 (mar 2017), \url{http://arxiv.org/abs/1703.01365}

\bibitem{toda2019convolutional}
Toda, Y., Okura, F., et~al.: How convolutional neural networks diagnose plant
  disease. Plant Phenomics  \textbf{2019},  9237136 (2019)

\bibitem{xie2020explainable}
Xie, N., Ras, G., van Gerven, M., Doran, D.: Explainable deep learning: A field
  guide for the uninitiated. arXiv preprint arXiv:2004.14545  (2020)

\bibitem{zoph2018learning}
Zoph, B., Vasudevan, V., Shlens, J., Le, Q.V.: {Learning transferable
  architectures for scalable image recognition}. In: Proceedings of the IEEE
  conference on computer vision and pattern recognition. pp. 8697--8710 (2018)

\end{thebibliography}
\end{document}